\documentclass[sigconf]{acmart}

\AtBeginDocument{%
  }

\usepackage{amsmath}
\usepackage{algorithmic}
\usepackage{algorithm}
\usepackage{multirow}
\usepackage{float}
\usepackage{graphicx}
\usepackage{subcaption}

\usepackage{tcolorbox}
\tcbuselibrary{breakable}

\usepackage{enumitem}

\usepackage[dvipsnames]{xcolor}
\usepackage{ifthen}
\newboolean{colorcomments}
\setboolean{colorcomments}{true} 

\newcommand{\bita}[1]{\ifthenelse{\boolean{colorcomments}}{\textcolor{Purple}{B: #1}}{#1}}

\setcopyright{acmlicensed}
\copyrightyear{2018}
\acmYear{2018}
\acmDOI{XXXXXXX.XXXXXXX}
\acmConference[Conference acronym 'XX]{Make sure to enter the correct
  conference title from your rights confirmation email}{June 03--05,
  2018}{Woodstock, NY}
\acmISBN{978-1-4503-XXXX-X/2018/06}

\begin{document}

\title{From Symbolic to Geometric: Enabling Spatial Reasoning in Large Language Models}

\author{Chen Chu}
\email{chenchu@usc.edu}
\affiliation{%
  \institution{University of Southern California}
  \city{Los Angeles}
  \state{CA}
  \country{USA}
}

\author{Bita Azarijoo}
\email{azarijoo@usc.edu}
\affiliation{%
  \institution{University of Southern California}
  \city{Los Angeles}
  \state{CA}
  \country{USA}
}

\author{Li Xiong}
\email{lxiong@emory.edu}
\affiliation{%
  \institution{Emory University}
  \city{Atlanta}
  \state{GA}
  \country{USA}
}

\author{Khurram Shafique}
\email{kshafique@novateur.ai}
\affiliation{%
 \institution{Novateur Research Solutions}
 \city{Ashburn}
 \state{Virginia}
 \country{USA}}

\author{Cyrus Shahabi}
\email{shahabi@usc.edu}
\affiliation{%
  \institution{University of Southern California}
  \city{Los Angeles}
  \state{CA}
  \country{USA}
}

\renewcommand{\shortauthors}{Trovato et al.}

\begin{abstract}

Recent large language models (LLMs) often appear to exhibit spatial reasoning ability; however, this capability is largely \emph{symbolic}, arising from pattern matching over spatial language rather than true \emph{geometric} reasoning over space. Because LLMs operate on discrete tokens, they lack native support for continuous spatial representations, explicit geometric computation, and structured spatial operators. To address this limitation, we introduce the \emph{Spatial Language Model (SLM)}, the first multimodal LLM that treats location information as a first-class modality and enables geometric spatial reasoning within the model's inference process. SLM directly operates on learned spatial representations rather than textual descriptions of spatial relations. To support effective training, we construct a \emph{Spatial Instruction Dataset} that aligns spatial representations, atomic geometric operations, and natural language instructions. We further propose a new benchmark named \emph{SpatialEval}, which is designed to evaluate spatial reasoning across attributes, distance, topology, and relative-position tasks. Extensive experiments show that SLM significantly outperforms existing LLM-based approaches that rely on symbolic reasoning via prompt engineering or textual abstraction, demonstrating the benefits of integrating geometric spatial representations for robust spatial reasoning.

Our instruction dataset, evaluation benchmark, model training codes, and models' checkpoints can be found at: 

\hyperlink{https://github.com/chuchen2017/SLM}{https://github.com/chuchen2017/SLM}.

\end{abstract}

\ccsdesc[500]{Computing methodologies~Artificial intelligence}
\keywords{Large Language Models, Multimodel Language Model, Spatial Reasoning, Spatial Representation Learning}

\received{20 February 2007}
\received[revised]{12 March 2009}
\received[accepted]{5 June 2009}

\maketitle

\section{Introduction}

Spatial reasoning is a fundamental capability for large language models (LLMs) to understand, reason about, and interact with the physical world \cite{spatialrgpt}. Such capability underpins a wide range of applications, including point-of-interest recommendation \cite{poi}, urban planning \cite{planning}, socioeconomic indicator prediction \cite{urbanvlp}, human mobility modeling \cite{geollama}, and geospatial knowledge extraction \cite{geollm}. Despite the impressive progress of modern LLMs, enabling robust and generalizable spatial reasoning remains an open challenge.

Recent studies have explored the extent to which LLMs can perform spatial reasoning using text-only inputs \cite{citygpt}. These works show that portions of an LLM's parameters implicitly encode geospatial relationships among named locations, learned from co-occurrence patterns in text corpora \cite{llmrepresentst}. As a result, LLMs can often reproduce or infer spatial relationships expressed linguistically, such as recalling that Paris lies in France. However, this capability primarily reflects \emph{symbolic spatial reasoning}; that is, reasoning over discrete linguistic symbols such as place names, coordinates, and relational terms, rather than \emph{geometric spatial reasoning} over explicit, continuous representations of space. When confronted with tasks that require structured spatial inference, such as topological reasoning or metric consistency, such as determining whether two given polygons contain each other, even state-of-the-art LLMs produce contradictory or logically invalid conclusions \cite{shaolin, song2025}.

The limitations of symbolic spatial reasoning in LLMs stem primarily from their reliance on language-level representations rather than explicit modeling of space. Most prior work evaluates spatial understanding through place names and textual descriptions rather than through explicit spatial representations \cite{gpt4geo}. Consequently, LLMs tend to answer spatial queries by retrieving or recombining previously observed linguistic patterns describing location relationships, rather than by reasoning over underlying geometry \cite{citygpt}. This pattern-retrieval mechanism is inherently brittle: it is prone to hallucination \cite{hallucination}, fails to generalize to newly introduced or unseen locations, and constrains spatial reasoning to entities explicitly encoded during training. These failures underscore a fundamental gap between symbolic reasoning over spatial language and genuine geometric reasoning over space.

In addition, symbolic spatial reasoning imposes intrinsic computational limitations. Many spatial tasks, including distance measurement, relative-position computation, require precise numerical and geometric operations. However, current LLMs lack reliable numerical computation capabilities \cite{nlpmath}, and geographic reasoning often involves large-magnitude coordinates and multi-step calculations. Consequently, performing accurate geometric spatial reasoning within LLMs remains fundamentally difficult.

As a result, recent approaches increasingly rely on prompt engineering, external tools, or agentic AI frameworks \cite{song2025,agent}. In such systems, spatial queries are decomposed into subtasks and resolved through calls to external services such as map APIs or web search engines. For example, Gemini invokes Google Maps to answer spatial queries, while ChatGPT relies on external web search to retrieve geospatial information. While these agentic solutions can improve empirical performance on specific tasks, they introduce new challenges: hallucinations may still occur, reasoning becomes heavily dependent on external knowledge sources, and inference is token- and latency-intensive due to tool invocation. Most importantly, these approaches treat geometric reasoning as an external service rather than an intrinsic model capability, and therefore do not integrate explicit geometric spatial representations into the LLM itself, leaving the core limitation of symbolic spatial reasoning fundamentally unaddressed.

Motivated by the limitations of symbolic spatial reasoning in current LLMs, we aim to equip language models with \emph{intrinsic geometric spatial reasoning} capabilities. In this work, we introduce the first spatial multimodal LLM, dubbed the \emph{Spatial Language Model (SLM)}, which treats geospatial representations as a first-class modality rather than encoding spatial relationships implicitly through language. By directly operating on learned spatial representations, SLM enables spatial reasoning within the model’s inference process, instead of relying on symbolic retrieval from textual training data. This design allows the model to generalize naturally to newly introduced and previously unseen spatial entities. Moreover, through intrinsic reasoning rather than numerical computation, the model achieves substantial improvements in spatial reasoning performance. Finally, by avoiding reliance on agentic tool invocation, SLM significantly reducing token consumption during inference.

The foundation of SLM is unified representation learning for heterogeneous geospatial entities. By mapping different types of spatial entities into a shared embedding space, SLM enables geometric reasoning across heterogeneous inputs using shared internal functions, rather than relying on entity-specific symbolic. This unified representation facilitates both efficient training and robust generalization. In this work, we adopt Geo2Vec \cite{geo2vec} with a adapter to learn such representations. Building on this foundation, we design a progressive training strategy that improves training efficiency. Finally, we introduce the \emph{SpatialEval}, which is the first spatial evaluation benchmark specifically designed to assess the both symbolic and geometric spatial reasoning capabilities of LLMs. 

A fundamental challenge in enabling geometric spatial reasoning in LLMs lies in the availability of suitable training data. Unlike vision–language or audio-based LLMs \cite{clip,onellm}, where data pairs can be easily obtained from interenet, there is no large-scale dataset that jointly provides precise geospatial representations and natural-language spatial instructions. To address this gap, we construct a \emph{Spatial Instruction Dataset} that decomposes complex spatial reasoning queries into a chain-of-thought style format. This dataset explicitly aligns spatial representations, atomic geometric operations, and language instructions, enabling LLMs to learn geometric spatial reasoning processes rather than symbolic pattern matching.

In summary, our contributions are as follows:
\begin{itemize}[leftmargin=16pt, topsep=0pt, itemsep=0pt]
\item[$\bullet$] We propose the first geospatial multimodal LLM that integrates intrinsic geometric spatial reasoning into language models.
\item[$\bullet$] We construct the first spatial instruction dataset that aligns spatial representations, geometric reasoning operations, and language instructions, supporting spatial reasoning across diverse input types.
\item[$\bullet$] We introduce a spatial reasoning evaluation benchmark designed to assess LLMs’ geometric spatial reasoning performance under heterogeneous spatial inputs.
\item[$\bullet$] Extensive empirical results demonstrate that SLM achieves significant improvements over existing LLM-based spatial reasoning approaches that rely on symbolic reasoning.
\end{itemize}

\section{Related Work}

\textbf{Multimodal Language Model}. Recent advances in LLMs have extended their strong reasoning capabilities to multimodal inputs. Early progress was driven by vision–language models \cite{clip}, followed by extensions to video \cite{video}, audio \cite{audio}, and more domain-specific modalities such as point clouds \cite{pointllm} and street-view imagery \cite{urbanllava}. These multimodal language models enable LLMs to jointly reason over linguistic and non-linguistic information, significantly broadening their applicability. Most multimodal LLMs follow a common architectural paradigm. A modality-specific encoder is first used to transform raw inputs into high-level feature representations. These features are then projected into the language model’s embedding space through an adapter or projection layer, allowing the LLM to align and integrate multimodal information with natural language \cite{llava}. To the best of our knowledge, existing multimodal LLMs have not been designed to explicitly support spatial reasoning over geometric representations. In this research, we present the first attempt to integrate explicit geometric spatial representations into the multimodal LLM framework, enabling intrinsic spatial reasoning that goes beyond symbolic, text-based spatial representations.


\smallskip \noindent
\textbf{Spatial Representation Learning}. Recent advances in Spatial Representation Learning (SRL) form a key foundation for building geometry-based language models. SRL aims to learn unified vector representations for diverse types of geospatial entities \cite{SRL}. Traditional methods typically support only a single type of geospatial entity \cite{NUFT}, which limits downstream models from performing unified spatial reasoning. More recently, Poly2Vec \cite{poly2vec} and Geo2Vec \cite{geo2vec} have been proposed to enable unified representation learning across heterogeneous geospatial entities. This unified learning paradigm provides downstream models with coherent and consistent spatial information, preventing them from learning separate functions for conceptually similar spatial reasoning tasks. By leveraging this property of unified representation learning, we are able to compress traditionally complex symbolic representations of geospatial entities into a single vector representation, enabling more efficient and effective spatial reasoning within language models.

\begin{figure*}[h]
    \centering
    \includegraphics[width=0.8\textwidth]{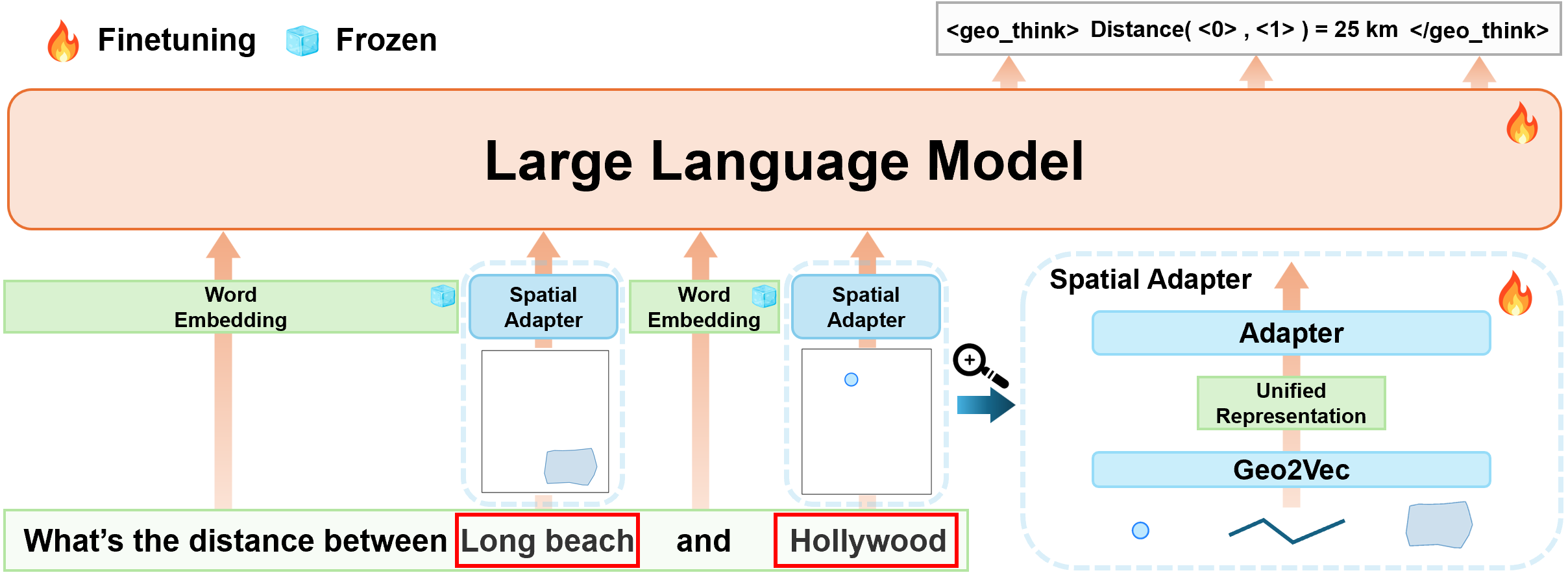} 
    \caption{Framework of the proposed Spatial Language Model.}
    \label{figure1}
    \Description{Framework of the proposed Spatial Language Model.}
\end{figure*}

\section{Methodology}

Figure~\ref{figure1} illustrates the overall framework of SLM. A spatial reasoning query is first parsed to identify interleaved geospatial entities, which are then converted into geometric representations and projected through a spatial adapter into the embedding space. These embeddings are integrated into the LLM, enabling intrinsic awareness of spatial information.

\subsection{Geometric Representation}

Geospatial data naturally exist in heterogeneous geometric forms. Enabling geometric spatial reasoning in LLMs requires operating over a unified spatial representation rather than relying on symbolic or entity-specific encodings. Accordingly, we introduce a unified spatial encoder that maps heterogeneous spatial entities into a shared embedding space, allowing the model to acquire a consistent geometric understanding of space. In this work, we adopt a state-of-the-art spatial representation learning method Geo2Vec \cite{geo2vec} to encode geospatial entities into vector representations that capture both geometric shape and absolute location. This method functions as a unified spatial encoder, supporting heterogeneous geospatial entities within a shared embedding space. As a result, SLM can perform geometric spatial reasoning consistently across different entity types. Formally, we define:
{\setlength{\abovedisplayskip}{2pt}
 \setlength{\belowdisplayskip}{2pt}
\begin{equation}
V_E = \mathcal{E}(E) 
,\end{equation}}
where, for any given geospatial entity \(E\) (e.g. POIs, roads, buildings, regions), the spatial encoder \(\mathcal{E}\) encodes it to a vector representation \(V_E \in R ^ {d}\), \(d\) is the number of dimension of a spatial representation. 

Subsequently, following the standard practice in building multimodal LLMs, we employ a spatial adapter $f_{\text{ada}}$ to project \(V_E\) to the pretrained word embedding space:
{\setlength{\abovedisplayskip}{2pt}
 \setlength{\belowdisplayskip}{2pt}
\begin{equation}
V'_E = f_{\text{ada}}(V_E)
,\end{equation}}
where \(V'_E \in R ^ {d'}\), and \(d'\) denotes the dimensionality of the language model’s word embedding space.

\subsection{Interleaved Geometric Grounding}

Geospatial entities appear in natural language in an interleaved form, which requires our SLM to be able to process such inputs effectively. To this end, we propose an interleaved prompt format. For any spatial entity in a prompt, it is represented in the form:
{\setlength{\abovedisplayskip}{2pt}
 \setlength{\belowdisplayskip}{2pt}
\[
\texttt{<NAME> <PHRASE> <GEO>}
\]}
where "\texttt{<NAME>}" denotes the entity identifier, which typically corresponds to the name of a spatial entity, such as a road name, POI name, city name, or just index. "\texttt{<PHRASE>}" corresponds to location description phrases, for example, “located at,” “at,” or it may be omitted. We further introduce a special token "\texttt{<GEO>}" which serves as a placeholder for the spatial modality input and will be replaced by the spatial representation vector \(V'_E\) corresponding to the given entity. For example, a query such as \emph{What's the area of Long Beach?} will be transformed into \emph{What's the area of \texttt{<NAME> <PHRASE> <GEO>}?}, where \texttt{<PHRASE>} will be taken over by "located at", and \texttt{<GEO>} will be replaced by the geometric representation of Long Beach during model processing. In this way, the downstream LLM is able to access and learn to reason directly over the explicit geometric representation of a geospatial entity.

When multiple geospatial entities are mentioned in a single prompt, we distinguish them by adding subscripts to the entity identifiers, for example, \texttt{<NAME\_0>} and \texttt{<NAME\_1>}. Since each spatial representation is associated with its corresponding identifier, no additional distinction is required. For the rest of the paper, for simplicity, we adopt a simplified representation of the proposed spatial entity prompt structure, using \texttt{<E\_0>}, \texttt{<E\_1>}, and so on to denote the specific entities appearing in a sentence.

Specifically, we aim to enable the model to perform spatially-aware inference directly on learned spatial embeddings, rather than relying on the retrieval of prior knowledge associated with textual place identifiers. Therefore, we do not provide the model with the exact names of locations. Instead, we use an index token in the form "\texttt{<INDEX>}" to replace "\texttt{<NAME>}", which assigns each place a unique index identifier within a given prompt and allows the model to distinguish among different entities.

Nevertheless, to support general purpose spatial LLM, spatial entities should be presented in this format using their actual names during pretraining or fine-tuning stage, so that spatial entities should ideally be presented in this interleaved format using their actual names, so that the model can learn the alignment between a place name and its corresponding spatial location. 

\subsection{Spatial Instruction Dataset Construction}

To effectively align spatial representations with textual embeddings, it is essential to construct a fine-tuning dataset that not only provides correct answers, but also exposes the underlying reasoning process required to derive them. We therefore design our \emph{Spatial Instruction Dataset} in a question answering format, where complex spatial reasoning questions are decomposed into atomic functions using a geospatial chain-of-thought representation. This design encourages SLM to learn explicit spatial operations within the model, providing clear supervision signals for spatial reasoning.

The formulation of the dataset follows established practices in GeoQA research, following previous practice in VLM's spatial reasoning dataset construction \cite{spatialrgpt}, in which spatial reasoning tasks are naturally expressed as factoid geographic questions, which contain three sub-categories: \textbf{single geographic entity questions}, \textbf{spatial relationship questions}, and \textbf{spatial qualifier questions} \cite{geoqa}. By structuring the data in this way, we ensure that the learning objective of the model is tightly coupled with real-world spatial reasoning demands.

\noindent
\underline{\textbf{Single Geographic Entity Questions}}.
This type of question focuses on the attributes of a single geospatial entity. It does not require computation involving multiple entities, as the reasoning process only involves deriving properties from one entity. In other words, it requires the model to extract attribute-level information from a geometric representation. In the \emph{Spatial Instruction Dataset}, we include questions such as computing the area of a polygon entity or the length of a polyline entity. For example: \emph{What is the area of <E\_0>?} or \emph{How long is <E\_0>?} For this category, the model’s response is supervised using an atomic functional form. Specifically, attribute-related answers follow the format:
{\setlength{\abovedisplayskip}{2pt}
 \setlength{\belowdisplayskip}{2pt}
\[
\begin{aligned}
&\texttt{<geo\_think>}\\
&\texttt{Area(<NAME>) = <ANSWER>}\\
&\texttt{</geo\_think>}
\end{aligned}
\]
}where \texttt{<NAME>} and \texttt{<ANSWER>} are placeholders corresponding to the previously defined entity identifier and the computed result of the atomic function, respectively.

\noindent
\underline{\textbf{Spatial Relationship Questions}}.
Spatial relationship questions focus on reasoning about the relative positions and interactions between two or more geospatial entities. These tasks require the model to infer relationships such as distance, containment, intersection. Typical examples like: \emph{What is the distance between <E\_0> and <E\_1>?}

In the \emph{Spatial Instruction Dataset}, we design this category to cover fundamental spatial relations, including metric relations (e.g., distance between two entities) and topological relations (e.g., contains, intersects, and within). This category therefore serves as a core component for enhancing the model’s ability to perform multi-entity spatial inference. For this type of question, the answer is defined as of an atomic function that specifies the involved entity identifiers. For example, for distance-based questions, the model is expected to generate:
{\setlength{\abovedisplayskip}{2pt}
 \setlength{\belowdisplayskip}{2pt}
\[
\begin{aligned}
&\texttt{<geo\_think>} \\
&\texttt{Distance(<NAME\_0>, <NAME\_1>) = <ANSWER>} \\
&\texttt{</geo\_think>}
\end{aligned}
\]}where \texttt{<NAME\_0>} and \texttt{<NAME\_0>} denote the identifiers of the two spatial entities, and \texttt{<ANSWER>} represents the computed distance between them. This formulation provides explicit supervision for learning spatial relationship, enabling the model to internalize spatial reasoning rather than relying on symoblic computing.

\noindent
\underline{\textbf{Spatial Qualifier Questions}}.
This class of questions refers to those that involve a set of geospatial entities subject to one or more spatial qualifiers. Such questions typically require comparative computations among multiple entities. For example: \emph{Given <E\_0>, <E\_1>, and <E\_2>, which one is closer to <E\_3>?}. The answer to this type of question is composed of the atomic functions defined earlier. For instance, the response to the example above is formulated as:
{\setlength{\abovedisplayskip}{2pt}
 \setlength{\belowdisplayskip}{2pt}
\[
\begin{aligned}
&\texttt{<geo\_think>} \\
&\texttt{Distance(<NAME\_0>, <NAME\_3>) = <ANSWER>} \\
&\texttt{Distance(<NAME\_1>, <NAME\_3>) = <ANSWER>} \\
&\texttt{Distance(<NAME\_2>, <NAME\_3>) = <ANSWER>} \\
&\texttt{</geo\_think>}
\end{aligned}
\]}This structured formulation enables the model to explicitly reason over multiple candidate entities through a sequence of fundamental spatial functions, supporting reliable comparison-based spatial inference rather than relying on model's intuition.

\noindent
\underline{\textbf{Prompt Evolution}}.
Diversity in training prompts is crucial for improving the generalization ability of the fine-tuned SLM. Relying solely on a limited set of manually designed questions may cause the model to overfit to specific linguistic patterns, thereby weakening its robustness to unseen query formulations. To address this issue, we adopt a prompt evolution strategy to systematically expand the dataset and increase the diversity, while preserving their underlying spatial reasoning intent \cite{selfinstruct}. 

Specifically, for each category of spatial reasoning queries, we first collect a small set of seed questions from human experts and online sources. These seed questions serve as canonical examples that clearly define the reasoning focus of each task category. We then leverage several strong LLMs, including Llama-3.3-70B and ChatGPT 5.2, to automatically generate new questions conditioned on these seeds. During this process, the LLMs are instructed to produce questions that maintain the same spatial reasoning intention, while exhibiting substantial diversity in linguistic structure, phrasing, and contextual framing. The generation instruction prompt can be found in the Appendix.

This prompt evolution process enables the construction of a large-scale and linguistically diverse \emph{Spatial Instruction Dataset} without requiring extensive manual annotation. More importantly, by exposing SLM to varied natural-language expressions of the same underlying spatial operations, the model is encouraged to learn invariant spatial reasoning patterns rather than memorizing, thereby significantly enhancing its generalization capability in open-world scenarios. After cleaning, there are in total 30k all types queries in the instruction dataset. 

\noindent
\underline{\textbf{Agent-guided Answer Construction}}. After the prompt evolution stage, many automatically generated questions no longer strictly follow the original structure, particularly for spatial qualifier questions. These questions often express complex spatial intentions that require the composition of multiple atomic spatial functions to arrive at a correct answer. In addition, solving such questions requires the model not only to perform accurate spatial computations within each atomic function, but also to correctly ground and reference the specific geospatial entities involved. Moreover, redundant or unnecessary computations should be avoided to ensure that the reasoning process remains concise and logically consistent.

To build reasonable responses for those questions, we employ an LLM agent to guide the construction of structured answers. Specifically, the agent is provided with a predefined set of atomic spatial functions that it is allowed to invoke. Given a spatial reasoning question, the agent is instructed to solve the problem by decomposing it into a sequence of atomic function calls, while explicitly grounding each operation to the involved geospatial entities using the previously defined structured format. For each query, the agent’s function invocation trajectory is recorded. This trajectory is then structured into the predefined \texttt{<geo\_think>} format and used as the final answer corresponding to the generated question.

This agent-guided answer construction strategy ensures that each query in the \emph{Spatial Instruction Dataset} is paired with an explicit reasoning trace that reflects the underlying spatial operations required to solve the task. By supervising the model with explicit, minimal, and correctly grounded atomic function sequences, we enable SLM to learn compositional spatial reasoning patterns, particularly for complex spatial qualifier queries, while reducing noise introduced by unstructured or redundant reasoning paths.

It is worth noting that, although our method leverages agent-generated data to provide structured supervision during training, it fundamentally differs from the agent-based inference approaches discussed earlier. By distilling structured reasoning into the model, we reduce reliance on symbolic retrieval and tool invocation. Beyond improving inference efficiency, it encourages the model to approximate geometric functions directly from entity representations, moving toward intrinsic spatial reasoning within the SLM.

\subsection{Spatial Evaluation Benchmark}

\begin{figure}
    \centering
    \includegraphics[width=0.7\columnwidth]{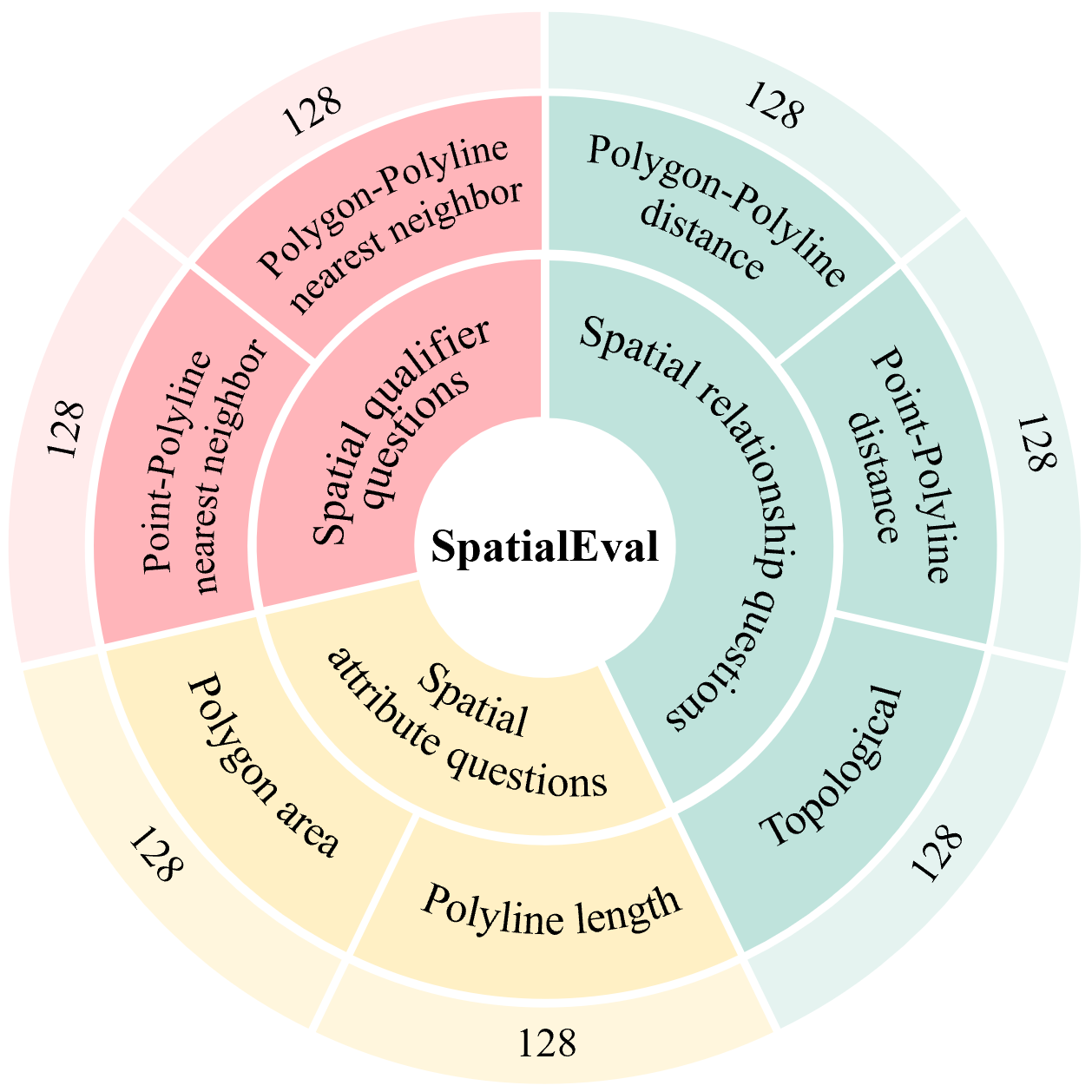}
    \caption{Composition of \emph{SpatialEval} benchmark.}
    \label{figure2}
    \Description{Questions composition of SpatialEval.}
\end{figure}

\begin{table*}[]
\caption{Performance of different LLMs on single entity questions in \emph{SpatialEval}. Area denotes polygon area estimation, and Length denotes polyline length estimation. Values in parentheses indicate the response validity rate of valid answers. $\downarrow$ indicates lower is better. \textbf{Best} results are bold. }
\label{table1}
\small
\begin{tabular}{cccccccc}
\hline
                             & \multicolumn{2}{c}{Beijing}                                                                                                        &  & \multicolumn{2}{c}{Los Angeles}                                                                                                     &  & US Boundaries                                                       \\ \cline{2-3} \cline{5-6} \cline{8-8} 
Models                       & Area $\downarrow$                                                            & Length $\downarrow$                                                          &  & Area $\downarrow$                                                              & Length $\downarrow$                                                          &  & Area $\downarrow$                                                            \\ \hline

Qwen3-8B                     & \begin{tabular}[c]{@{}c@{}}4093.5\\ (100.)\end{tabular}      & \begin{tabular}[c]{@{}c@{}}2324.0\\ (97.7)\end{tabular}        &  & \begin{tabular}[c]{@{}c@{}}5053.4\\ (97.7)\end{tabular}      & \begin{tabular}[c]{@{}c@{}}2534.2\\ (91.4)\end{tabular}       &  & \begin{tabular}[c]{@{}c@{}}663.679\\ (100.)\end{tabular}     \\

Qwen3-8B (tuned)                     & \begin{tabular}[c]{@{}c@{}}1809.8\\ (100.)\end{tabular}      & \begin{tabular}[c]{@{}c@{}}1034.0\\ (100.)\end{tabular}        &  & \begin{tabular}[c]{@{}c@{}}1687.1\\ (100.)\end{tabular}      & \begin{tabular}[c]{@{}c@{}}573.1\\ (100.)\end{tabular}       &  & \begin{tabular}[c]{@{}c@{}}0.210\\ (100.)\end{tabular}     \\

DeepSeek-R1-32B & \begin{tabular}[c]{@{}c@{}}5293.7\\ (13.3)\end{tabular}           & \begin{tabular}[c]{@{}c@{}}888.0\\ (58.6)\end{tabular}          &  & \begin{tabular}[c]{@{}c@{}}8979.6\\ (37.5)\end{tabular}           & \begin{tabular}[c]{@{}c@{}}417.2\\ (37.5)\end{tabular}        &  & \begin{tabular}[c]{@{}c@{}}193.921\\ (9.4)\end{tabular}             \\

Llama-3.3-70B       & \begin{tabular}[c]{@{}c@{}}115291.6\\ (91.7)\end{tabular}      & \begin{tabular}[c]{@{}c@{}}1093.7\\ (100.)\end{tabular}        &  & \begin{tabular}[c]{@{}c@{}}815900.9\\ (82.8)\end{tabular}       & \begin{tabular}[c]{@{}c@{}}1260.9\\ (79.7)\end{tabular}        &  & \begin{tabular}[c]{@{}c@{}}1490.118\\ (75.0)\end{tabular}        \\ \hline

gemini-2.5-flash             & \textbf{\begin{tabular}[c]{@{}c@{}}1217.3\\ (100.)\end{tabular}}          & \begin{tabular}[c]{@{}c@{}}309.3\\ (100.)\end{tabular}         &  & \begin{tabular}[c]{@{}c@{}}1562.4\\ (100.)\end{tabular}           & \textbf{\begin{tabular}[c]{@{}c@{}}156.4\\ (100.)\end{tabular}   }       &  & \begin{tabular}[c]{@{}c@{}}223.059\\ (100.)\end{tabular}          \\

ChatGPT 5.1                      & \begin{tabular}[c]{@{}c@{}}1314.9\\ (100.)\end{tabular} & \begin{tabular}[c]{@{}c@{}}610.2\\ (100.)\end{tabular}         &  & \begin{tabular}[c]{@{}c@{}}1291.6\\ (100.)\end{tabular}          & \begin{tabular}[c]{@{}c@{}}279.4\\ (100.)\end{tabular}         &  & \begin{tabular}[c]{@{}c@{}}2803.246\\ (100.)\end{tabular}          \\

 ChatGPT 5.1 In-Context                     & \begin{tabular}[c]{@{}c@{}}1314.9\\ (100.)\end{tabular} & \begin{tabular}[c]{@{}c@{}}610.2\\ (100.)\end{tabular}         &  & \begin{tabular}[c]{@{}c@{}}1137.8\\ (100.)\end{tabular}          & \begin{tabular}[c]{@{}c@{}}279.4\\ (100.)\end{tabular}         &  & \begin{tabular}[c]{@{}c@{}}574.45\\ (100.)\end{tabular}          \\

 \hline
 
SpatialRGPT                          & \begin{tabular}[c]{@{}c@{}}2092.62\\ (100.)\end{tabular}         & \begin{tabular}[c]{@{}c@{}}263.35\\ (100.)\end{tabular} &  & \begin{tabular}[c]{@{}c@{}}1698.8\\ (100.)\end{tabular} & \begin{tabular}[c]{@{}c@{}}274.8\\ (100.)\end{tabular} &  & \begin{tabular}[c]{@{}c@{}}0.148\\ (100.)\end{tabular} \\

 \hline
 
SLM                          & \begin{tabular}[c]{@{}c@{}}1376.0\\ (100.)\end{tabular}         & \textbf{\begin{tabular}[c]{@{}c@{}}231.0\\ (100.)\end{tabular} }&  & \textbf{\begin{tabular}[c]{@{}c@{}}1175.3\\ (100.)\end{tabular}} & \begin{tabular}[c]{@{}c@{}}203.8\\ (100.)\end{tabular} &  & \textbf{\begin{tabular}[c]{@{}c@{}}0.082\\ (100.)\end{tabular}} \\

SRL                  & 1050.9                                                          & 60.6                                                           &  & 884.4                                                            & 38.9                                                           &  & 0.055                                                           \\ \hline
\end{tabular}
\end{table*}

Existing spatial reasoning benchmarks for LLMs suffer from limitations in input modality. Most datasets either do not include the actual geographic coordinates of spatial entities or rely solely on indirect textual descriptions of locations. Benchmarks based on symbolic representation primarily evaluate a model’s ability to recall memorized geographic facts from its training corpus, rather than its true spatial reasoning capacity. The lack of multimodal support makes it difficult to comprehensively assess the effectiveness of spatial reasoning methods that operate through alternative pipelines, such as coordinate or representation based methods.

To enable a more principled and flexible evaluation, we argue that a spatial reasoning benchmark should explicitly include both the spatial reasoning queries and the corresponding geographic coordinates of the spatial entities mentioned in each query. Such a design preserves the ability to evaluate both symbolic reasoning and direct geometric reasoning, thereby enabling assessment of intrinsic spatial reasoning capabilities.

Based on this motivation, we design a new spatial reasoning benchmark using the proposed spatial entity grounding method. By replacing the geospatial entity placeholder with different input modalities, such as entity names, coordinate representations, or vector embeddings, the benchmark supports direct evaluation of LLMs under various geospatial input settings.

The benchmark covers a broad range of spatial operations, including metric reasoning (e.g., distance estimation), geometric reasoning (e.g., area computation), and topological reasoning (e.g., topological classification). A set of evaluation questions is provided to ensure fair and reproducible comparison across models. We named our benchmark as \emph{SpatialEval} and the composition of the benchmark can be found in Figure \ref{figure2}. Samples can be found in Appendix.

\section{Experiments}

\subsection{Experiment Settings}

\noindent\textbf{Dataset Preparation}.
It is worth noting that the \emph{Spatial Instruction Dataset} we construct consists solely of spatial reasoning questions paired with their corresponding structured answer templates, it does not contain grounded geospatial entities. This design allows the instruction dataset to be flexibly instantiated with different underlying geospatial datasets. When adapting the instruction dataset to a specific geospatial dataset, the entity placeholders are replaced with concrete spatial entities sampled from the target dataset.

To test the effectiveness of such a design, we conduct experiments on three datasets across two spatial scales: points of interest (POIs), road networks, and building footprints from Beijing and Los Angeles at the urban scale (Beijing and Los Angeles datasets), as well as polygonal boundaries of counties and states in the United States at the national scale (US Boundaries dataset). Each dataset is split into training and testing subsets. Instruction prompts are filled with entities sampled from the training set. All benchmark queries are constructed exclusively from the testing sets. Dataset statistics can be found in the Appendix. 

\noindent\textbf{Implementation Details}.
For SLM, we select Qwen3-8B \cite{qwen3} as the base model for fine-tuning. Fine-tuning is performed using LoRA \cite{lora}. Notably, during training, SLM is trained only base on geospatial representations from the training dataset. This setting ensures that, during benchmark evaluation, the model performs spatial reasoning based on the provided geospatial embeddings rather than relying on memorizing all the embeddings. More model setting details could be found in the Appendix. 

\noindent\textbf{Baselines}.
We select a diverse set of LLMs with different parameter scales as baselines, including both open-source and proprietary models. The open-source LLMs consist of Qwen3-8B \cite{qwen3}, DeepSeek-R1-Distill-Qwen-32B \cite{Deepseek}, and LLaMA-3.3-70B-Instruct \cite{llama3}, while proprietary LLMs include Gemini 2.5 Flash \cite{gemini25} and ChatGPT 5.1 \cite{chatgpt}. Since these baseline LLMs only support text modality, we follow prior work on symbolic spatial reasoning in LLMs \cite{song2025} by providing the \textbf{explicit geographic coordinates} of each entity as their spatial representation. The detailed query prompts used for evaluation are provided in the Appendix.

We also introduced a VLM in comparison. Specifically, we adopted the spatial representation framework from SpatialRGPT \cite{spatialrgpt}, which couples a Vision Transformer with a specialized mask pooling module. This integration allows the model to mitigating the spatial degradation common in standard flattened VLM architectures. We finetuned the modal on our instruction dataset to suit it to the specific geospatial domains. 

We do not include agentic methods as accuracy baselines, as our goal is to evaluate the intrinsic spatial reasoning capabilities of LLMs rather than their ability to decompose problems using external tools. However, agentic approaches are included in our analysis of computational efficiency for comparison.

Due to the complexity of certain spatial reasoning queries, some baseline models fail to produce valid answers. In these cases, models may exhibit degenerate behaviors such as repetitive outputs, irrelevant reasoning, or explicit refusals indicating inability to solve the task. We treat such responses as failures. Accordingly, in addition to task-specific performance metrics, we report the response validity rate of each model, defined as the proportion of queries for which the model produces a valid and complete answer.

Since SLM learns spatial reasoning based on the provided spatial representations, its performance is inherently influenced by the quality of these representations. To quantify this effect, we additionally report results obtained by directly applying a multi-layer perceptron (MLP) to the spatial representations for task-specific inference. The performance serves as an approximate upper bound for SLM. We denote this baseline as SRL (Spatial Representation Learning) and report its results in the experimental section.

\begin{table*}[]
\small
\caption{Performance of different LLMs on entity relational questions in SpatialEval. PT, PL, and PG denote point, polyline, and polygon entities, respectively. For example, PT–PL denotes a relational question between point and polyline. Values in parentheses indicate the response validity rate. $\downarrow$ indicates lower is better. $\uparrow$ indicates higher is better. \textbf{Best} results are bold. }
\label{table2}
\begin{tabular}{ccccccccccccc}
\hline
                 & \multicolumn{4}{c}{Beijing}                                                                                                                                                                                                                                             &           & \multicolumn{4}{c}{Los Angeles}                                                                                                                                                                                                                                           &           & \multicolumn{2}{c}{US Boundaries}                                                                                                   \\ \cline{2-5} \cline{7-10} \cline{12-13} 
                 & \multicolumn{2}{c}{Distance $\downarrow$ }                                                                                                      & \multicolumn{2}{c}{NN $\uparrow$}                                                                                                              &           & \multicolumn{2}{c}{Distance $\downarrow$ }                                                                                                        & \multicolumn{2}{c}{NN $\uparrow$}                                                                                                              &           & Distance $\downarrow$                                                       & Topo $\uparrow$                                                              \\ \cline{2-5} \cline{7-10} \cline{12-13} 
                 & PT-PL                                                           & PG-PL                                                           & PT-PL                                                            & PG-PL                                                            &           & PT-PL                                                            & PG-PL                                                            & PT-PL                                                            & PG-PL                                                            &           & PG-PG                                                         & PG-PG                                                           \\ \hline
Qwen3-8B   & \begin{tabular}[c]{@{}c@{}}7816.0\\ (52.3)\end{tabular}     & \begin{tabular}[c]{@{}c@{}}5881.3\\ (6.3)\end{tabular}        & \begin{tabular}[c]{@{}c@{}}12.50\%\\ (29.7)\end{tabular}                                                               & \begin{tabular}[c]{@{}c@{}}0.78\%\\ (1.6)\end{tabular}                                                               &           & \begin{tabular}[c]{@{}c@{}}10152.1\\ (40.6)\end{tabular}     & \begin{tabular}[c]{@{}c@{}}9054.9\\ (15.6)\end{tabular}       & \begin{tabular}[c]{@{}c@{}}21.88\%\\ (49.2)\end{tabular}                                                               & \begin{tabular}[c]{@{}c@{}}3.91\%\\ (11.7)\end{tabular}                                                               &           & \begin{tabular}[c]{@{}c@{}}57.43\\ (10.9)\end{tabular}       & \begin{tabular}[c]{@{}c@{}}50.0\%\\ (100.)\end{tabular}           \\

Qwen3-8B (tuned)   & \begin{tabular}[c]{@{}c@{}}6392.0\\ (100.)\end{tabular}     & \begin{tabular}[c]{@{}c@{}}6390.1\\ (100.)\end{tabular}        & \begin{tabular}[c]{@{}c@{}}20.31\%\\ (100.)\end{tabular}                                                               & \begin{tabular}[c]{@{}c@{}}28.91\%\\ (100.)\end{tabular}                                                               &           & \begin{tabular}[c]{@{}c@{}}8862.2\\ (100.)\end{tabular}     & \begin{tabular}[c]{@{}c@{}}8440.9\\ (100.)\end{tabular}       & \begin{tabular}[c]{@{}c@{}}21.88\%\\ (100.)\end{tabular}                                                               & \begin{tabular}[c]{@{}c@{}}28.12\%\\ (100.)\end{tabular}                                                               &           & \begin{tabular}[c]{@{}c@{}}13.37\\ (100.)\end{tabular}       & \begin{tabular}[c]{@{}c@{}}53.1\%\\ (100.)\end{tabular}           \\

DeepSeek-R1-32B  & \begin{tabular}[c]{@{}c@{}}4972.8\\ (3.1)\end{tabular}        & \begin{tabular}[c]{@{}c@{}}3980.4\\ (37.5)\end{tabular}          & \begin{tabular}[c]{@{}c@{}}10.2\%\\ (10.9)\end{tabular}                                                                        & \begin{tabular}[c]{@{}c@{}}60.9\%\\ (63.3)\end{tabular}                                                               &           & \begin{tabular}[c]{@{}c@{}}7029.7\\ (1.0)\end{tabular}        & \begin{tabular}[c]{@{}c@{}}5209.5\\ (46.9)\end{tabular}         &\begin{tabular}[c]{@{}c@{}}13.3\%\\ (14.1)\end{tabular}                                                               & \begin{tabular}[c]{@{}c@{}}69.5\%\\ (78.1)\end{tabular}                                                               &           & \begin{tabular}[c]{@{}c@{}}5080.49\\ (42.2)\end{tabular}           & \begin{tabular}[c]{@{}c@{}}67.2\%\\ (100.)\end{tabular}           \\

Llama-3.3-70B    & \begin{tabular}[c]{@{}c@{}}32121.7\\ (98.4)\end{tabular}      & \begin{tabular}[c]{@{}c@{}}33496.4\\ (100.)\end{tabular}      & \begin{tabular}[c]{@{}c@{}}57.0\%\\ (100.)\end{tabular}                                                                & \begin{tabular}[c]{@{}c@{}}48.4\%\\ (100.)\end{tabular}                                                                &           & \begin{tabular}[c]{@{}c@{}}50821.1\\ (100.)\end{tabular}         & \begin{tabular}[c]{@{}c@{}}43225.1\\ (100.)\end{tabular}         & \begin{tabular}[c]{@{}c@{}}57.0\%\\ (100.)\end{tabular}                                                                   & \begin{tabular}[c]{@{}c@{}}52.3\%\\ (100.)\end{tabular}                                                                    &           & \begin{tabular}[c]{@{}c@{}}106.56\\ (100.)\end{tabular}        & \begin{tabular}[c]{@{}c@{}}68.8\%\\ (100.)\end{tabular}           \\ \hline

gemini-2.5-flash w/ name & \begin{tabular}[c]{@{}c@{}}10057.1\\ (88.3)\end{tabular}         & \begin{tabular}[c]{@{}c@{}}23940.6\\ (85.9)\end{tabular}         & \begin{tabular}[c]{@{}c@{}}22.65\%\\ (84.4)\end{tabular}          & \begin{tabular}[c]{@{}c@{}}21.1\%\\ (78.9)\end{tabular}           &           & \begin{tabular}[c]{@{}c@{}}90384.3\\ (53.9)\end{tabular}          & \begin{tabular}[c]{@{}c@{}}15177.7\\ (97.7)\end{tabular}          & \begin{tabular}[c]{@{}c@{}}21.9\%\\ (83.6)\end{tabular}          & \begin{tabular}[c]{@{}c@{}}20.3\%\\ (76.6)\end{tabular}          &           & \begin{tabular}[c]{@{}c@{}}31195.93\\ (32.8)\end{tabular}         & \begin{tabular}[c]{@{}c@{}}39.1\%\\ (93.0)\end{tabular}       \\

GPT 5.1 w/ name         & \begin{tabular}[c]{@{}c@{}}6346.1\\ (4.6)\end{tabular}         & \begin{tabular}[c]{@{}c@{}}6095.7\\ (0.1)\end{tabular}         & \begin{tabular}[c]{@{}c@{}}30.46\%\\ (88.3)\end{tabular}          & \begin{tabular}[c]{@{}c@{}}25.0\%\\ (93.0)\end{tabular}          &           & \begin{tabular}[c]{@{}c@{}}6628.7\\ (0.)\end{tabular}          & \begin{tabular}[c]{@{}c@{}}7791.0\\ (0.7)\end{tabular}          & \begin{tabular}[c]{@{}c@{}}22.7\%\\ (73.4)\end{tabular}          & \begin{tabular}[c]{@{}c@{}}21.9\%\\ (79.7)\end{tabular}          &           & \begin{tabular}[c]{@{}c@{}}55.23\\ (3.1)\end{tabular}        & \begin{tabular}[c]{@{}c@{}}39.1\%\\ (89.9)\end{tabular}  \\

\hline

gemini-2.5-flash & \begin{tabular}[c]{@{}c@{}}2600.1\\ (100.)\end{tabular}         & \begin{tabular}[c]{@{}c@{}}1665.5\\ (100.)\end{tabular}         & \begin{tabular}[c]{@{}c@{}}71.9\%\\ (100.)\end{tabular}          & \begin{tabular}[c]{@{}c@{}}75.0\%\\ (100.)\end{tabular}           &           & \begin{tabular}[c]{@{}c@{}}1687.7\\ (100.)\end{tabular}          & \begin{tabular}[c]{@{}c@{}}3021.1\\ (100.)\end{tabular}          & \begin{tabular}[c]{@{}c@{}}76.6\%\\ (100.)\end{tabular}          & \begin{tabular}[c]{@{}c@{}}74.2\%\\ (100.)\end{tabular}          &           & \begin{tabular}[c]{@{}c@{}}22.47\\ (100.)\end{tabular}         & \begin{tabular}[c]{@{}c@{}}33.6\%\\ (100.)\end{tabular}       \\

ChatGPT 5.1          & \begin{tabular}[c]{@{}c@{}}23023.7\\ (100.)\end{tabular}         & \begin{tabular}[c]{@{}c@{}}7039.6\\ (100.)\end{tabular}         & \begin{tabular}[c]{@{}c@{}}66.4\%\\ (100.)\end{tabular}          & \begin{tabular}[c]{@{}c@{}}68.8\%\\ (100.)\end{tabular}          &           & \begin{tabular}[c]{@{}c@{}}4697.4\\ (100.)\end{tabular}          & \begin{tabular}[c]{@{}c@{}}8572.2\\ (100.)\end{tabular}          & \begin{tabular}[c]{@{}c@{}}70.3\%\\ (100.)\end{tabular}          & \begin{tabular}[c]{@{}c@{}}71.1\%\\ (100.)\end{tabular}          &           & \begin{tabular}[c]{@{}c@{}}393.80\\ (100.)\end{tabular}        & \begin{tabular}[c]{@{}c@{}}49.2\%\\ (100.)\end{tabular}     \\

ChatGPT 5.1 In-Context         & \begin{tabular}[c]{@{}c@{}}9571.7\\ (100.)\end{tabular}         & \begin{tabular}[c]{@{}c@{}}6458.3\\ (100.)\end{tabular}         & \begin{tabular}[c]{@{}c@{}}67.2\%\\ (100.)\end{tabular}          & \begin{tabular}[c]{@{}c@{}}70.3\%\\ (100.)\end{tabular}          &           & \begin{tabular}[c]{@{}c@{}}9527.7\\ (100.)\end{tabular}          & \begin{tabular}[c]{@{}c@{}}8116.5\\ (100.)\end{tabular}          & \begin{tabular}[c]{@{}c@{}}73.4\%\\ (100.)\end{tabular}          & \begin{tabular}[c]{@{}c@{}}75.8\%\\ (100.)\end{tabular}          &           & \begin{tabular}[c]{@{}c@{}}1376.11\\ (100.)\end{tabular}        & \begin{tabular}[c]{@{}c@{}}51.6\%\\ (100.)\end{tabular}     \\

\hline

SpatialRGPT              & \begin{tabular}[c]{@{}c@{}}4955.4\\ (100.)\end{tabular} & \begin{tabular}[c]{@{}c@{}}5534.5\\ (100.)\end{tabular} & \begin{tabular}[c]{@{}c@{}}33.6\%\\ (100.)\end{tabular} & \begin{tabular}[c]{@{}c@{}}28.1\%\\ (100.)\end{tabular} & \textbf{} & \begin{tabular}[c]{@{}c@{}}7290.3\\ (100.)\end{tabular} & \begin{tabular}[c]{@{}c@{}}8626.3\\ (100.)\end{tabular} & \begin{tabular}[c]{@{}c@{}}21.9\%\\ (100.)\end{tabular} & \begin{tabular}[c]{@{}c@{}}21.1\%\\ (100.)\end{tabular} & \textbf{} & \begin{tabular}[c]{@{}c@{}}6.49\\ (100.)\end{tabular} & \begin{tabular}[c]{@{}c@{}}55.5\%\\ (100.)\end{tabular} \\

\hline

SLM              & \textbf{\begin{tabular}[c]{@{}c@{}}784.9\\ (100.)\end{tabular}} & \textbf{\begin{tabular}[c]{@{}c@{}}589.0\\ (100.)\end{tabular}} & \textbf{\begin{tabular}[c]{@{}c@{}}80.5\%\\ (100.)\end{tabular}} & \textbf{\begin{tabular}[c]{@{}c@{}}85.9\%\\ (100.)\end{tabular}} & \textbf{} & \textbf{\begin{tabular}[c]{@{}c@{}}882.9\\ (100.)\end{tabular}} & \textbf{\begin{tabular}[c]{@{}c@{}}836.6\\ (100.)\end{tabular}} & \textbf{\begin{tabular}[c]{@{}c@{}}83.6\%\\ (100.)\end{tabular}} & \textbf{\begin{tabular}[c]{@{}c@{}}84.4\%\\ (100.)\end{tabular}} & \textbf{} & \textbf{\begin{tabular}[c]{@{}c@{}}0.82\\ (100.)\end{tabular}} & \textbf{\begin{tabular}[c]{@{}c@{}}87.2\%\\ (100.)\end{tabular}} \\

SRL            & 393.5                                                          & 349.3                                                          & 95.7\%                                                               &96.3\%                                                               &           & 543.6                                                           & 576.3                                                           & 95.1\%                                                               & 95.9\%                                                                    &           & 0.37                                                         & 97.3\%                                                         \\ \hline
\end{tabular}
\end{table*}

\subsection{Overall Performance on SpatialEval}

We report the evaluation results on \emph{SpatialEval} in Table~\ref{table1} and Table~\ref{table2}. Table~\ref{table1} summarizes performance on single-entity reasoning tasks, including polygon area estimation (Area) and polyline length estimation (Length). These tasks are evaluated using Mean Absolute Error (MAE). Table~\ref{table2} presents results on spatial relational reasoning tasks involving two or more geospatial entities. For nearest neighbor query (NN) and topological relationship classification (Topo) questions, performance is evaluated using accuracy, while distance estimation tasks are evaluated using MAE.

As shown in Tables, the open-source baseline models fail on most spatial reasoning tasks, especially for pure numeric tasks, like Area and Distance estimation. We observe that Qwen3 and Llama tend to answer these questions without explicit reasoning processes, which results in relatively high response validity rates but poor overall performance. This indicates that their intuitive spatial reasoning ability is weak when facing symbolic spatial representations. In contrast, the reasoning-oriented model DeepSeek exhibits relatively better accuracy, but its response validity rate is notably low. We find that the model attempts to solve queries through explicit mathematical computation, which requires a large number of tokens and can take several minutes to resolve even one simple query. During the process, the model often becomes overly focused on intermediate calculations rather than the original query. Moreover, even simple spatial attribute queries involve a sequence of numerical operations, during which accumulated errors can easily occur and propagate to the final result. These observations suggest that symbolic reasoning over raw coordinate representations is ineffective, both in terms of computational efficiency and inference accuracy.

We further fine-tuned a Qwen3-8B model using our spatial instruction dataset to enable it to directly answer geospatial reasoning queries and to investigate whether symbolic representations possess intrinsic spatial reasoning capabilities. After fine-tuning, the model achieved a 100\% response rate across all question categories, indicating that it successfully learned to follow the task format and generate valid answers. While its performance improved modestly compared to the untuned model, it remained substantially inferior to geometry-based methods. These results suggest that, despite benefiting from instruction tuning, symbolic representations alone have inherent limitations in capturing precise spatial relationships and supporting accurate geospatial reasoning. This highlights the fundamental advantage of explicitly incorporating geometric information rather than relying solely on symbolic descriptions of geospatial entities.

Proprietary LLMs also tend to answer directly without explicit reasoning, but exhibit stronger spatial intuition than open-source ones. This intuition performs well on relatively simple tasks such as Area and Length estimation, where the required limited spatial reasoning. Meanwhile, intuition-based answering breaks down when confronted with more complex relational queries, such as Distance estimation, Topological classification, and nearest-neighbor query. These tasks require the model to jointly reason over the relative locations of multiple spatial entities. These results show that, when facing symbolic spatial representations, even proprietary LLMs struggle to maintain coherent spatial reasoning, leading to degraded performance on complex spatial reasoning tasks. 

It is worth noting that the VLM-based baseline, SpatialRGPT, also struggles with relational reasoning tasks, such as distance estimation and topological relationship inference. However, it achieves relatively strong performance on attribute inference tasks, such as line-length estimation. This observation suggests that VLMs primarily rely on the positional encoding mechanisms of Vision Transformers to capture spatial relationships. While effective in many vision tasks, this approach may be suboptimal for geospatial scenarios, where target entities often occupy only a small fraction of the entire image. The patch-based spatial partitioning of ViTs can make it challenging for the model to accurately learn and represent the absolute positions of entities, thereby limiting its ability to perform fine-grained spatial reasoning and relational inference. VLMs incur significantly higher training costs because each entity must be represented by an image of the entire region, resulting in thousands of visual tokens after ViT encoding. In contrast, our SLM uses a single token to directly represent the entity's precise location, providing a much more efficient and scalable representation for geospatial reasoning.

SLM demonstrates consistently strong performance across all spatial reasoning tasks. For attribute questions, although the model’s accuracy is bounded by the quality of the underlying spatial representations, SLM still outperforms SOTA baselines on several tasks. For complex spatial reasoning tasks, SLM consistently surpasses all baseline models. Especially in distance estimation, where SLM achieves significantly better results than all symbolic-based approaches. Another notable strength of SLM is its stable performance across different geospatial entity types. Benefiting from the unified geometric vector representation, SLM is able to reason over heterogeneous entities. In contrast, symbolic-based models exhibit highly non-uniform performance when handling different entity types. For example, DeepSeek shows strong discrepancies between Point-Polyline and Polygon-Polyline reasoning. These results highlight the importance of unified geometric representations for robust and generalizable spatial reasoning.

We also evaluate proprietary LLMs using place names as symbolic inputs, instead of coordinates, with the corresponding results denoted as \textbf{with / name} in the tables. Although city information is provided to the models, both performance and response validity rates remain low. These results indicate that LLMs have limited exposure to fine-grained real-world geospatial data during training. This finding further highlights the poor generalization of purely textual symbolic representations for spatial reasoning.

\subsection{Spatial Generalization}

\begin{figure}
    \centering
    \begin{subfigure}[b]{0.48\columnwidth}
        \includegraphics[width=\linewidth]{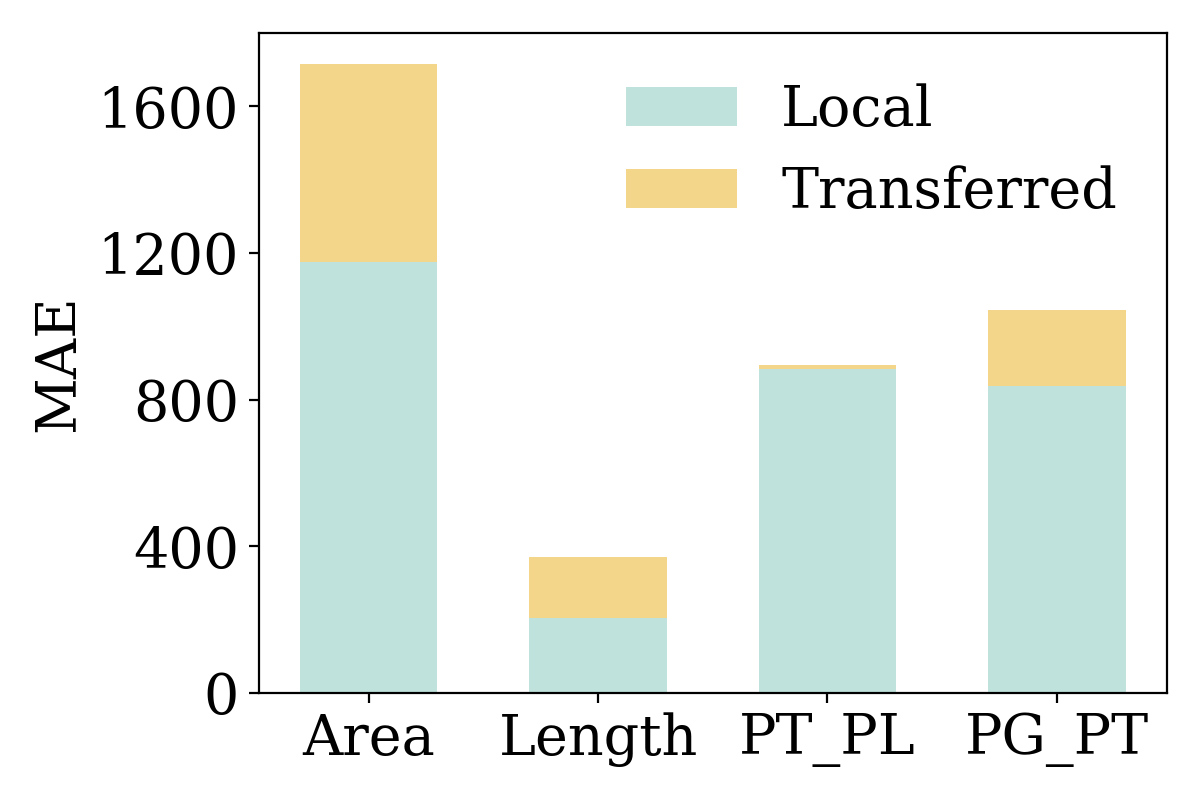}
        \caption{Los Angeles to Beijing \(\downarrow\)}
        \label{fig:sub1}
    \end{subfigure}
    \hfill
    \begin{subfigure}[b]{0.48\columnwidth}
        \includegraphics[width=\linewidth]{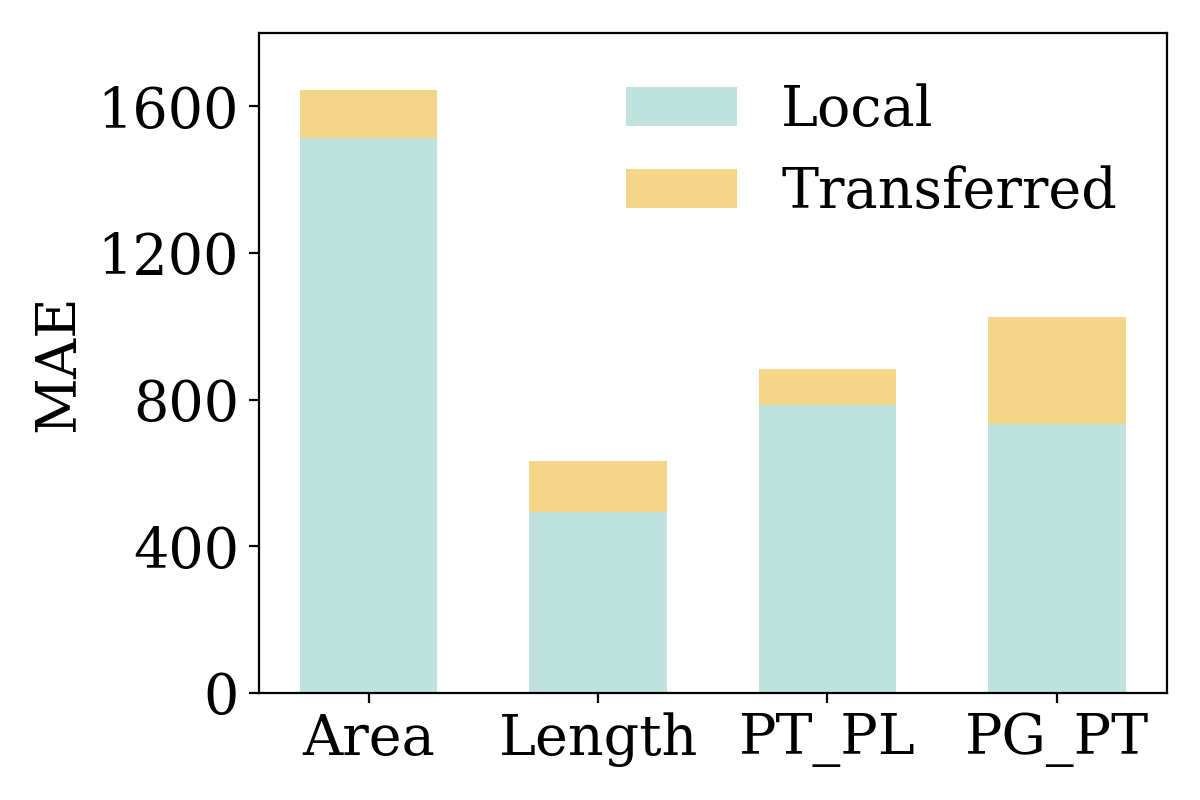}
        \caption{Beijing to Los Angeles \(\downarrow\)}
        \label{fig:sub2}
    \end{subfigure}
    \caption{Zero-shot inference performance of SLM on the target dataset (transferred) compared with models trained on the target dataset (local).}
    \label{figure3}
\end{figure}

SLM generalizes effectively to unseen geographic regions due to the transferability of its unified geometric representation learning framework. To evaluate this capability, we perform zero-shot inference on a target city using a model trained on a different city, with the results shown in Figure~\ref{figure3}. The largest degradation occurs in Area estimation, with a 46.0\% drop when transferring from Los Angeles to Beijing. We attribute this degradation to limitations in the underlying spatial representations, as area estimation remains a challenging task for current SRL methods. In contrast, SLM exhibits stable performance on all other tasks under zero-shot transfer, with the smallest observed decrease being only 0.01\%. These results suggest that the model learns robust and transferable spatial reasoning patterns that generalize across geographic regions. Based on this strong generalization ability, SLM can be trained on data-rich regions and subsequently deployed as a foundation model for spatial reasoning across a wide range of geographic datasets.

\subsection{Query Generalization}

\begin{figure}
\centering
\includegraphics[width=0.6\columnwidth]{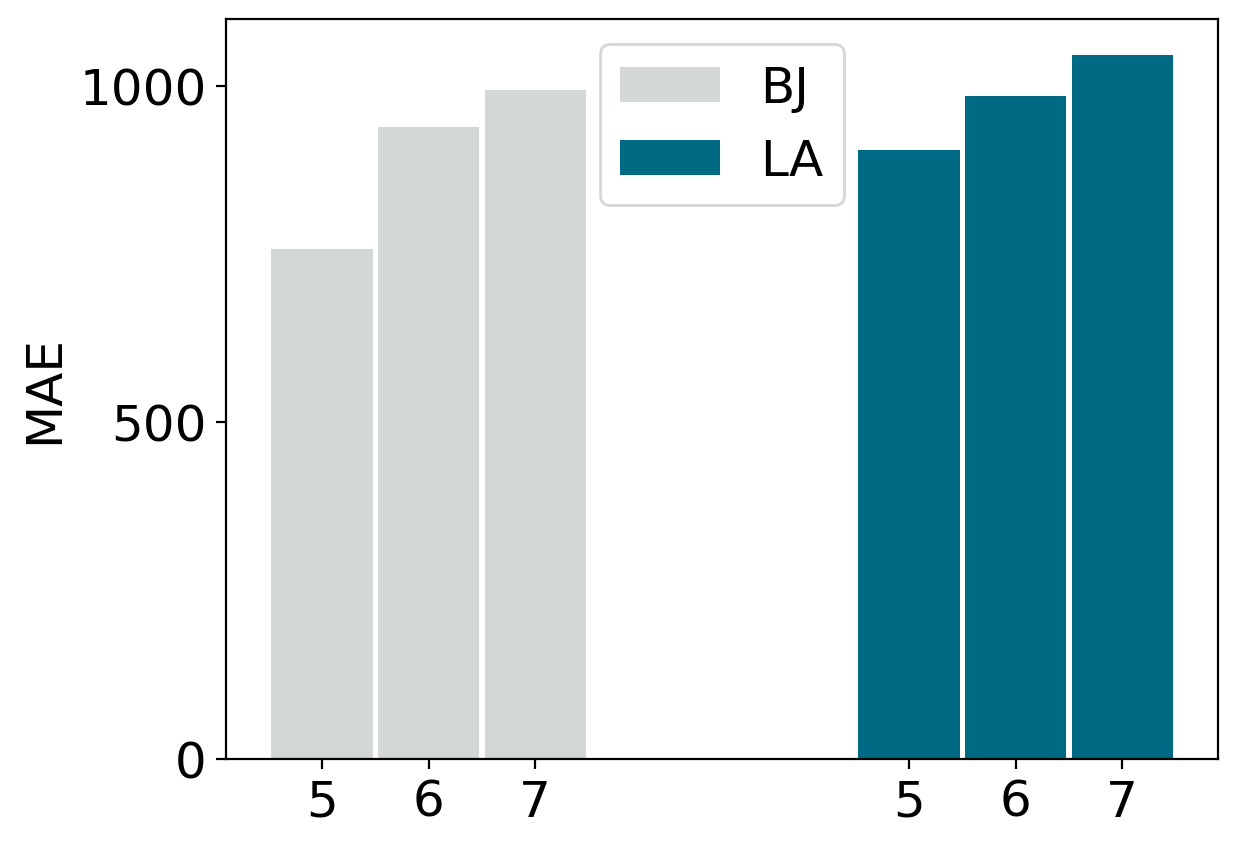}
\caption{SLM performance on spatial reasoning queries not included in the finetuning dataset. The X-Axis denotes the number of entities involved in the test queries; queries with 6 and 7 entities are unseen during training.}
\label{figure4}
\Description{Query Generalization.}
\end{figure}

Training on the \emph{Spatial Instruction Dataset} enables SLM to generalize beyond the specific query structures observed during training. In the instruction dataset, we deliberately restrict the number of involved geospatial entities to at most 5. In this section, we evaluate the model’s generalization ability by testing it on point–polyline nearest-neighbor queries involving 6 and 7 entities, which are not present in the training data. We assess model performance by computing the MAE for each prediction function. The results are shown in Figure~\ref{figure4}. Despite the increased query complexity and the larger number of involved entities, SLM remains capable of solving these unseen spatial reasoning queries. Although a slight performance degradation is observed, the model maintains stable and acceptable accuracy across both datasets, demonstrating strong query-level generalization beyond the training distribution.

\subsection{Computational Efficiency}

\begin{table}
\small
\caption{Comparison of Input token consumption, Output token consumption and inference Time among symbolic methods, agentic methods, and SLM. The values after ± indicate the standard-deviation.}
\label{table3}
\begin{tabular}{llll}
\hline
                                                              & Input          & Output          & Time /s          \\ \hline
\begin{tabular}[c]{@{}c@{}}Symbolic w/Reasoning\end{tabular} & 1078.6±273.6 & 3954.1±448.7 & 139.6±15.8 \\
Symbolic                                                      & 1078.6±273.6 & 39.9±50.4    & \textbf{1.2±1.2}  \\
Agentic                                                       & 1438.0±113.3   & 997.4±150.5   & 22.1±6.9   \\
SLM                                                           & \textbf{30±0.}          & \textbf{30.7±0.4}     & 1.9±0.1    \\ \hline
\end{tabular}
\end{table}

To compare the computational efficiency of different spatial reasoning approaches, we evaluate their token consumption and inference latency on the point–polyline distance estimation task. Since SLM is fine-tuned based on Qwen3-8B, we use the same backbone model to answer identical queries. In addition, we include an agentic solution that invokes an external distance estimation function using place names to perform inference.

As shown in Table~\ref{table3}, SLM demonstrates substantially better computational efficiency than both symbolic and agentic baselines. In terms of token usage, SLM requires only a constant and minimal number of input and output tokens, whereas symbolic reasoning methods incur significantly higher token consumption. Although symbolic methods without reasoning traces achieve low inference latency, its response validity rate is low.

The agentic approach reduce output token usage but still incur substantial input overhead due to tool invocation and external context exchange. Although an LLM can be fine-tuned to operate within a specific agentic function and may therefore be effective for certain spatial queries, such functionality is fundamentally orthogonal to the intrinsic spatial reasoning capability of SLM. In particular, the ability to invoke external tools does not imply internal spatial understanding. Conversely, if a model can be fine-tuned to invoke external tools for spatial inference, it can also be fine-tuned to perform spatial reasoning directly within the model, as demonstrated by SLM. Our results suggest that empowering LLMs with intrinsic spatial reasoning capabilities leads to a more efficient and stable solution than reliance on external agentic pipelines.

\subsection{Convergence Analysis}
\begin{figure}[ht]
    \centering
    \begin{subfigure}[b]{0.48\columnwidth}
        \includegraphics[width=\linewidth]{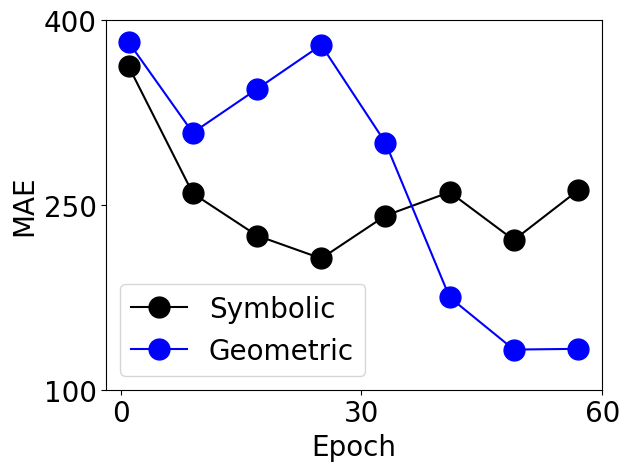}
        \caption{Line length inference \(\downarrow\)}
        \label{fig:sub1}
    \end{subfigure}
    \hfill
    \begin{subfigure}[b]{0.48\columnwidth}
        \includegraphics[width=\linewidth]{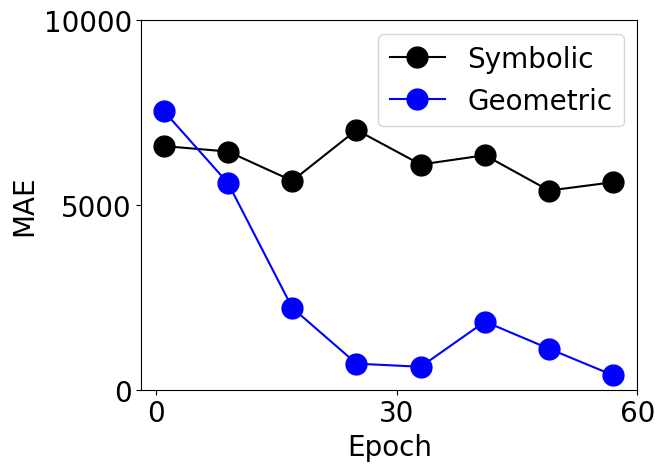}
        \caption{Distance estimation \(\downarrow\)}
        \label{fig:sub2}
    \end{subfigure}
    \caption{Performance comparison between symbolic (Qwen3-8B) and geometric (SLM) finetuning methods across different training epochs on the evaluation dataset.}
    \label{Figure 3}
\end{figure}

To further understand the learning dynamics of our proposed SLM compared to traditional symbolic approaches, we analyze their convergence behaviors during the fine-tuning process.
Figure~\ref{Figure 3} illustrates the Mean Absolute Error (MAE) across 60 training epochs for both the geometric method (SLM) and the symbolic method (Qwen3-8B fine-tuned on explicit geographic coordinates in text format) on two representative spatial reasoning tasks: line length inference and distance estimation.

As shown in Figure~\ref{fig:sub1}, the symbolic model initially shows a rapid decrease in MAE during the first 25 epochs. However, the error rate begins to increase and fluctuate in subsequent epochs. This instability suggests that the model struggles to generalize and may be overfitting to the linguistic patterns of symbolic coordinates rather than reasoning over them. In contrast, while our geometric SLM exhibits some initial fluctuation during the early stages of fine-tuning, it finally demonstrates a sharp, steady decline in MAE. By epoch 50, the geometric model converges to a significantly lower and stable error rate (approximately 130 MAE), highlighting its capability to learn and reason over spatial attributes.

The disparity in learning dynamics is even more visible in the distance estimation task, as depicted in Figure~\ref{fig:sub2}. Because this task requires the model to jointly reason over the relative locations of multiple spatial entities, it poses a significant challenge for language-level representations to reason over spatial entities. Consequently, the symbolic fine-tuning method fails to converge meaningfully throughout the entire 60-epoch fine-tuning period.

\section{Conclusions}

In this paper, we present SLM, the first multimodal language model designed for spatial reasoning. Through the proposed interleaved spatial grounding structure, we construct the first \emph{Spatial Instruction Dataset} that enables LLMs to acquire intrinsic spatial reasoning capabilities through unified spatial representation learning. In addition, we introduce \emph{SpatialEval}, the first benchmark designed to assess spatial reasoning capacity under diverse geospatial input representations, including both symbolic and geometric forms.

Extensive experiments show that even state of the art proprietary LLMs exhibit limited spatial reasoning performance when relying on symbolic representations, such as entity coordinates or place names. In contrast, when provided with geometric representations, SLM achieves consistently strong performance across of spatial reasoning tasks. In sum, SLM achieves inference efficiency comparable to symbolic methods without reasoning traces, while surpassing the accuracy of symbolic approaches that rely on explicit reasoning, and does so in a manner orthogonal to agentic, tool-based pipelines.

Overall, our findings highlight intrinsic spatial reasoning as a fundamental yet underexplored capability for future LLMs. Unlike agentic approaches that depend on external tools and incur additional inference overhead, SLM integrates spatial reasoning directly within the model, offering a complementary and scalable direction for advancing spatial intelligence in large language models. In future work, we plan to investigate how intrinsic spatial understanding can help LLMs better organize their semantic space and further improve performance on general GeoQA tasks.

\begin{acks}
This research has been funded in part by the NIH awards R01LM014026, NSF award CNS-2125530 and DMS-2428039, and the Intelligence Advanced Research Projects Activity (IARPA) via the Department of Interior/Interior Business Center (DOI/IBC) contract number 140D0423C0033. The U.S. Government is authorized to reproduce and distribute reprints for the Governmental purposes, notwithstanding any copyright annotation thereon. Disclaimer: The views and conclusions contained herein are those of the authors and should not be interpreted as necessarily representing the official policies or endorsements, either expressed or implied, of NIH, NSF, IARPA, or the U.S. Government. In addition, this research project has benefited from the unrestricted cash gifts from Google Research.
\end{acks}

\bibliographystyle{ACM-Reference-Format}
\bibliography{sample-base}

\clearpage   
\appendix

\section*{Appendix}
\addcontentsline{toc}{section}{Appendix}

\section{Experiment details}

All experiments and model training were conducted on a single compute node equipped with 8 NVIDIA A6000 Ada GPUs. All open-source models were deployed locally and evaluated using their default inference parameters.

Fine-tuning was performed using LoRA, with a learning rate of $5\times10^{-5}$. The LoRA hyperparameters were set to $\alpha=64$ and $r=32$. This setup introduces 1.05\% additional trainable parameters. The dimension $d$ of geometric representation is set as 256 for Beijing and Los Angeles datasets, and 512 for US states dataset. Finetuning a model for one dataset takes around 8 hours to achieve the reported performance. 

\section{Dataset Statistics}

We present statistics of the \emph{Spatial Instruction Dataset} in Table~\ref{table4}, which reports the number of questions in each category. In addition, Table~\ref{table5} summarizes the dataset statistics, showing the number of each type of geospatial entity in the evaluated datasets.

\begin{table}[h]
\caption{Composition of the \emph{Spatial Instruction Dataset}}
\label{table4}
\begin{tabular}{cc}
\hline
Types            & Number \\ \hline
Area             & 5000   \\
Length           & 5000   \\
Distance         & 5000   \\
Topo             & 5000   \\
Nearest Neighbor & 10000  \\ \hline
\end{tabular}
\end{table}

\begin{table}[h]
\caption{Geospatial dataset statistics}
\label{table5}
\begin{tabular}{ccccc}
\hline
              & Point & Polyline & Polygon & Multipolygon \\ \hline
Beijing       & 3505  & 4562     & 7942    & 0            \\
Los Angeles   & 6914  & 10976    & 6436    & 0            \\
US Boundaries & 0     & 0        & 0       & 3158         \\ \hline
\end{tabular}
\end{table}

\section{Spatial Instruction Dataset Prompts}

In this section, we present several representative seed prompts used in the construction of the Spatial Instruction Dataset. We also provide the prompt templates employed during the prompt evolution process to generate diverse spatial reasoning queries. The special token \texttt{<E\_0>} will be replaced by the interleaved geometric grounding representation \texttt{<NAME> <PHRASE> <GEO>}.

\subsection{Instruction Dataset Seed Prompt Demos}

\begin{tcolorbox}[
  title=Prompt,
  colback=gray!5,
  colframe=black!60,
  fonttitle=\bfseries,
  breakable
]

- How long is the road \texttt{<E\_0>}?

- How long is the polyline segment \texttt{<E\_0>}?

- What's the length of the road \texttt{<E\_0>}?

- I am now walking on road \texttt{<E\_0>}; how long does the road last?

- The length of \texttt{<E\_0>} is:

- How large does \texttt{<E\_0>} cover?

- What is the area of the polygon \texttt{<E\_0>}?

- Can you tell me the size of the area enclosed by \texttt{<E\_0>}?

- I am interested in the extent of \texttt{<E\_0>}; what is its area?

- Given \texttt{<E\_0>}, what is the total area it covers?

- The area of \texttt{<E\_0>} is:

- What's the distance between \texttt{<E\_0>} and \texttt{<E\_1>}?

- Given \texttt{<E\_0>} and \texttt{<E\_1>}, what is the distance between them?

- Among \texttt{<E\_0>} and \texttt{<E\_1>}, how far apart are they?

- What is the length of the straight line connecting \texttt{<E\_0>} and \texttt{<E\_1>}?

- Can you tell me the distance from \texttt{<E\_0>} to \texttt{<E\_1>}?

- The distance between \texttt{<E\_0>} and \texttt{<E\_1>} is:

- Distance(\texttt{<E\_0>}, \texttt{<E\_1>}) =

- What's the distance between \texttt{<E\_0>} and \texttt{<E\_1>}?

- Given \texttt{<E\_0>} and \texttt{<E\_1>}, which one has a shorter distance to \texttt{<E\_2>}?

- Between \texttt{<E\_0>} and \texttt{<E\_1>}, which is closer to \texttt{<E\_2>}?

- Which location among \texttt{<E\_0>}, \texttt{<E\_1>}, and \texttt{<E\_2>} is farthest from \texttt{<E\_3>}?

- Between \texttt{<E\_0>} and \texttt{<E\_1>}, how are they related spatially?

- Between \texttt{<E\_0>} and \texttt{<E\_1>}, what is the topological relationship?

- Between \texttt{<E\_0>} and \texttt{<E\_1>}, can you describe the spatial connection in topological terms?

- Given \texttt{<E\_0>} and \texttt{<E\_1>}, what is their topological relationship?

- Between \texttt{<E\_0>} and \texttt{<E\_1>}, explain their spatial relationship.

- Does \texttt{<E\_0>} intersect with \texttt{<E\_1>}?

- Does \texttt{<E\_0>} contain \texttt{<E\_1>}?

- Are \texttt{<E\_0>} and \texttt{<E\_1>} disjoint?

- Does \texttt{<E\_0>} touch \texttt{<E\_1>}?

- Are \texttt{<E\_0>} in \texttt{<E\_1>}?

- Are \texttt{<E\_0>} within \texttt{<E\_1>}?

- Between \texttt{<E\_0>}, \texttt{<E\_1>}, which one is the closet to \texttt{<E\_2>}?

- Among \texttt{<E\_0>}, \texttt{<E\_1>}, and \texttt{<E\_2>}, which is closest to \texttt{<E\_3>}?

- Given \texttt{<E\_0>}, \texttt{<E\_1>}, \texttt{<E\_2>}, which one is the closet to \texttt{<E\_3>}?

- Given \texttt{<E\_0>}, \texttt{<E\_1>}, \texttt{<E\_2>}, which one is the farthest to \texttt{<E\_3>}?

- From \texttt{<E\_0>}, \texttt{<E\_1>}, \texttt{<E\_2>}, which deviates most from \texttt{<E\_3>}?

- Which is closer to \texttt{<E\_3>}: \texttt{<E\_0>}, \texttt{<E\_1>}, or \texttt{<E\_2>}?

- Between \texttt{<E\_0>} and \texttt{<E\_1>}, which is farther from \texttt{<E\_2>}?

- Among \texttt{<E\_0>}, \texttt{<E\_1>}, and \texttt{<E\_2>}, which is farthest from \texttt{<E\_3>}?

- Which of \texttt{<E\_0>} or \texttt{<E\_1>} is least similar to \texttt{<E\_2>}?

- From \texttt{<E\_0>}, \texttt{<E\_1>}, \texttt{<E\_2>}, which deviates most from \texttt{<E\_3>}?

\end{tcolorbox}

\subsection{Prompt Template for Evolution}

The token \texttt{<QUESTION>} will be replaced by one of the seed prompts.

\begin{tcolorbox}[
  title=Prompt,
  colback=gray!5,
  colframe=black!60,
  fonttitle=\bfseries,
  breakable
]

\textbf{System:}  
You are a helpful assistant focusing on spatial reasoning.

\vspace{0.5em}
\textbf{Instruction:}  
You \textbf{must} keep all special tokens in the form \texttt{<E\_number>} unchanged and preserve their original structure.

\vspace{0.5em}
\textbf{User:}  
Provide five questions in Python list format that have similar meaning to the initial question: \texttt{<QUESTION>}. The new questions should differ in phrasing while preserving the original intent, and should focus on either the attribute of a specific entity or the relationship between entities represented by \texttt{<E\_number>}.

\end{tcolorbox}

\section{Query Prompt List}

Here we present the prompt used to query the LLMs. The placeholder token \texttt{<GEO>} is replaced by the corresponding spatial modality input—such as a symbolic representation, a place name, or a geometric vector embedding—during model inference. The \texttt{<DATASET>} token is instantiated with the name of the target dataset (e.g., Beijing, Los Angeles, or the United States).
\subsection{Polygon Area Prediction}

\begin{tcolorbox}[
  title=Prompt,
  colback=gray!5,
  colframe=black!60,
  fonttitle=\bfseries,
  breakable
]

\textbf{System:}  
You are a helpful assistant specialized in spatial reasoning. You are at <DATASET>.

\vspace{0.5em}
\textbf{Instruction:}  
Given a polygon description, compute its area in the same units as the input coordinates.

Answer in the following format:
\begin{verbatim}
<answer> 12.34 </answer>
\end{verbatim}

Output only a single numeric value and do not include units.  

If the answer is unknown, return:
\begin{verbatim}
<answer> unknown </answer>
\end{verbatim}

\vspace{0.5em}
\textbf{User:}  
What is the area of the polygon \texttt{<0>} covered by \texttt{<GEO>}?

\end{tcolorbox}

\subsection{Polyline length Prediction}

\begin{tcolorbox}[
  title=Prompt,
  colback=gray!5,
  colframe=black!60,
  fonttitle=\bfseries,
  breakable
]

\textbf{System:}  
You are a helpful assistant specialized in spatial reasoning. You are at <DATASET>.

\vspace{0.5em}
\textbf{Instruction:}  
Given a polyline description, compute its length in the same units as the input coordinates.

Answer in the following format:
\begin{verbatim}
<answer> 12.34 </answer>
\end{verbatim}

Output only a single numeric value and do not include units.  

If the answer is unknown, return:
\begin{verbatim}
<answer> unknown </answer>
\end{verbatim}

\vspace{0.5em}
\textbf{User:}  
What's the length of \texttt{<0>} represented by \texttt{<GEO>}?

\end{tcolorbox}

\subsection{Distance Estimation}

\begin{tcolorbox}[
  title=Prompt,
  colback=gray!5,
  colframe=black!60,
  fonttitle=\bfseries,
  breakable
]

\textbf{System:}  
You are a helpful assistant specialized in spatial reasoning. You are at <DATASET>.

\vspace{0.5em}
\textbf{Instruction:}  
Given two geospatial entity descriptions, it could be point, polyline or polygon, compute the distance between two entities in the same units as the input coordinates.

Answer in the following format:
\begin{verbatim}
<answer> 12.34 </answer>
\end{verbatim}

Output only a single numeric value and do not include units.  

If the answer is unknown, return:
\begin{verbatim}
<answer> unknown </answer>
\end{verbatim}

\vspace{0.5em}
\textbf{User:}  
What is the distance between \texttt{<0>} located at \texttt{<GEO>} and \texttt{<1>} located at \texttt{<GEO>}?

\end{tcolorbox}

\subsection{Nearest Neighbor Query}

\begin{tcolorbox}[
  title=Prompt,
  colback=gray!5,
  colframe=black!60,
  fonttitle=\bfseries,
  breakable
]

\textbf{System:}  
You are a helpful assistant specialized in spatial reasoning. You are at \texttt{<DATASET>}.

\vspace{0.5em}
\textbf{Instruction:}  
Given a set of geospatial entities, which may be points, polylines, or polygons, find the nearest neighbor of the queried entity.

Answer in the following format:
\begin{verbatim}
<answer> <1> </answer>
\end{verbatim}

Output the index of the nearest entity.  

If the answer is unknown, return:
\begin{verbatim}
<answer> unknown </answer>
\end{verbatim}

\vspace{0.5em}
\textbf{User:}  
Given \texttt{<0>} located at \texttt{<GEO>}, \texttt{<1>} located at \texttt{<GEO>}, \texttt{<2>} located at \texttt{<GEO>}, and \texttt{<3>} located at \texttt{<GEO>}, which one is closest to \texttt{<4>} located at \texttt{<GEO>}?

\end{tcolorbox}

\end{document}